\documentclass[wcp]{jmlr}

\usepackage{longtable}

\usepackage{booktabs}
\usepackage{microtype}
\usepackage{graphicx}
\usepackage{subfloat}
\usepackage{booktabs}
\usepackage{mathrsfs}
\usepackage{multirow}
\usepackage{algorithm}
\usepackage{algorithmic}
\usepackage{skmath}
\usepackage{setspace}
\usepackage{booktabs} 

\usepackage{hyperref}

\newcommand{\ie}{\textit{i}.\textit{e}.}

\makeatletter
\def\set@curr@file#1{\def\@curr@file{#1}} 
\makeatother
\usepackage[load-configurations=version-1]{siunitx} 

\jmlrvolume{129}
\jmlryear{2020}
\jmlrworkshop{ACML 2020}

\title[DDBF]{Deep Dynamic Boosted Forest}

\author{\Name{Haixin Wang}\thanks{These authors have contributed to this work equally. 
 (\textit{Corresponding author: J.Sun.\textsuperscript{$\dagger$}})
 } \Email{wang.hx@pku.edu.cn}\\
 \Name{Xingzhang Ren}\footnotemark[1] \Email{xzhren@pku.edu.cn}\\
 \Name{Jinan Sun}\textsuperscript{$\dagger$} \Email{sjn@pku.edu.cn}\\
 \Name{Wei Ye} \Email{wye@pku.edu.cn}\\
 \Name{Long Chen} \Email{clcmlxl@pku.edu.cn}\\
 \Name{Muzhi Yu} \Email{muzhi.yu@pku.edu.cn}\\
 \Name{Shikun Zhang} \Email{zhangsk@pku.edu.cn}\\
 \addr Peking University, Beijing, China
}

\begin{document}

\maketitle

\begin{abstract}
 Random forest is widely exploited as an ensemble learning method. In many practical applications, however, there is still a significant challenge to learn from imbalanced data. To alleviate this limitation, we propose a deep dynamic boosted forest (DDBF), a novel ensemble algorithm that incorporates the notion of hard example mining into random forest. Speciﬁcally, we propose to measure the quality of each leaf node of every decision tree in the random forest to determine hard examples. By iteratively training and then removing easy examples from training data, we evolve the random forest to focus on hard examples dynamically so as to balance the proportion of samples and learn decision boundaries better. Data can be cascaded through these random forests learned in each iteration in sequence to generate more accurate predictions. Our DDBF outperforms random forest on 5 UCI datasets, MNIST and SATIMAGE, and achieved state-of-the-art results compared to other deep models. Moreover, we show that DDBF is also a new way of sampling and can be very useful and efficient when learning from imbalanced data.
\end{abstract}

\section{Introduction}
\label{introduction}

\begin{figure}[t]
    \centering
    \includegraphics[width=0.45\textwidth]{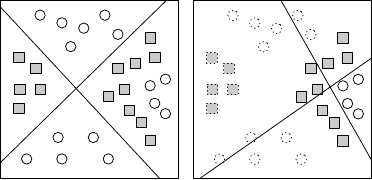} 
    \caption{Illustration of removing easy examples to improve learning. When we train a classifier as shown on the left, one of the decision regions is not ideal, but when we remove the easy examples from the good decision region and then train a classifier as shown in the right figure. At this point, the ensemble of the two classifiers allows the training data to be perfectly divided.}
    \label{fig:1}
\end{figure}
Recently, deep learning has achieved great success in various applications especially in classification tasks. However, neural networks still remain a black box as of today, which makes the training severely dependent on hyper-parameters tuning. Therefore, some researchers start to combine deep learning with traditional ensemble learning methods such as Random Forest (RF). In \cite{frosst2017distilling,kontschieder2015deep}, RF is exploited to enhance interpretability and performance with deep models. In turn, there's also advancement in deep forest \cite{zhou1702deep} with representation learning
\cite{feng2018autoencoder}.

Two of the most popular ensemble algorithms are bagging and boosting. RF \cite{breiman2001random} is the typical paradigm of bagging algorithms. Research on other advanced ensemble learning methods also have been established early. It has been found in \cite{kotsiantis2004combining} that boosting algorithms are stronger than bagging, but bagging is more robust in noisy settings. To utilize both merits, bagging and boosting are proposed to combine together for better performance. Besides, with the development of deep forests, boosted cascading structure has been further shown to be powerful against missing and imbalanced data in classification \cite{jones2001rapid}. 

From two aspects above, we observe several limitations of RF in development: 1) It can only be extended vertically (more decision trees) but not horizontally since the decision trees exist in parallel and cannot be stacked in layers in the same fashion as neurons in networks; 2) These decision trees have the same weight in voting for the final prediction despite that some of these trees may perform poorly; 3) all points in training data have the same weight and are treated equally in the sampling and training process, although some of them are easy to classify while others are hard. 

\begin{figure*}
    \centering
    \noindent\makebox[\textwidth]{%
    \includegraphics[width=0.85\textwidth]{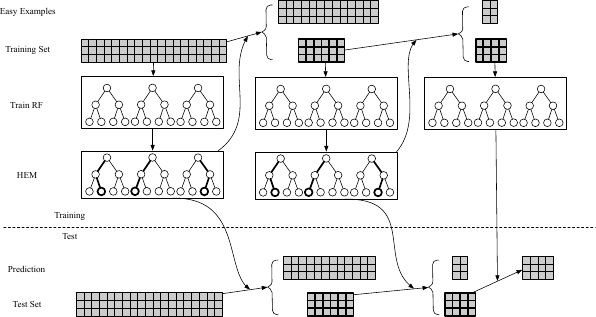}}
    \caption{Illustration of DDBF with three iterations. During training (above the dotted line), firstly, a random forest is trained on the training set, and then the leaf nodes are evaluated through the HEM process (Section \ref{hem}). The hard examples (with bold borders) can be identified accordingly (by removing easy examples) as a new training set for the next iteration. Similarly, at each iteration during test (below the dotted line), easy examples are identified (except the last iteration) and their predictions are outputted.}
    \label{fig:2}
\end{figure*}

In this paper, to alleviate the limitation of imbalanced data, we propose deep dynamic boosted forest (DDBF), a decision tree ensemble approach incorporating boosted cascaded structure into deep forest to overcome these limitations. It is a novel ensemble algorithm where such notion of hard examples mining as in boosting is incorporated into RF shown in Section \ref{2}. By means of iteratively training and then removing easy examples to train again, we evolve the boosted cascaded model to deep forests to focus on hard examples dynamically so as to learn decision boundaries better. A key point of our method is the definition of hard examples and easy ones, which we score the quality of each leaf node in the forest to vote for. It gives rise to the focus on data near the decision boundaries for better performance especially in imbalanced distributed samples. After training, the test data can be cascaded through these learned random forests of each iteration in sequence to generate predictions. In addition, we propose to design smart evolution mechanism and iteration mechanism to enhance the performance of DDBF and conducted ablation test. We evaluate DDBF on public datasets and it outperforms RF and achieves state-of-the-art results compared to other deep models. Last but not least, we also perform visualizations to validate DDBF and analyze its effectiveness from a sampling point of view.

The contributions of this paper is listed as follows:
\begin{itemize}
\item We propose Deep Dynamic Boosted Forest (DDBF), a novel ensemble algorithm that combine boosting and bagging into deep forests by means of iteratively mining and training on hard examples. 
\item We propose an evolution mechanism which filters out poor decision trees in the forest, together with iteration mechanism to enhance the performance of DDBF and achieve state-of-the-art results on both structured datasets and unstructured datasets.
\item The proposed hard example mining process can be seen as a novel way of sampling. We demonstrate the effectiveness of using DDBF to deal with imbalanced data.
\end{itemize}

\section{Motivation}
\label{2}
In typical classification tasks, classifiers are trained to find the decision boundaries of labelled data using a set of features. The distribution of data near the decision boundaries largely affect the performance of the classifier and is generally where the over-fitting and under-fitting trade-off is made. Those data whose predicted labels are not agreed upon across multiple classifiers, probably located near the decision boundaries, can be seen hard examples. On the contrary, those data whose predicted labels are consistent and correct across all classifiers can be seen as easy examples. Naturally, the performance on hard examples determines how good a classifier is compared with others and its ideal to focus on these hard examples during training.

Hard example mining is a common notion in may machine learning algorithms. A typical example is AdaBoost, in which wrongly classified examples are deemed as hard examples. The general idea is to assign more weight to those hard examples and train the model iteratively until convergence. This hard example mining notion has also been adopted in deep learning such as the boosted cascading structure \cite{chen2008fast}.

Thus, we are motivated to incorporate this hard example mining notion, or boosted cascading structure, into deep forest. It can be integrated smoothly within the RF training process since a random forest is an ensemble of decision trees trained on randomly sampled data and features, which are weak classifiers. If all decision trees make predictions reaching our criterion on a part of the training data, these data can be considered as easy examples. Rather than assign more weight to hard examples, we can simply remove these data from training set instead. When training the model in the next iteration, the model can focus on hard examples to learn decision boundaries better without adding extra parameters and complexity as shown in Figure \ref{fig:1}.

 One point to clarify is that the aforementioned criterion to determine easy examples, \ie data that are classified right across multiple classifiers, is not strict enough for removal because in extreme cases where data are imbalanced, we may be left with data of only one label. We will explain our proposed criterion in detail in Section \ref{propose}. But basically, with RF, we are seeking to find good rules and define easy examples as ones that fit the good rules of all classifiers. In a decision tree, each leaf node represents a rule, thus we need to come up with a criterion to determine whether a leaf node is good or not.

\begin{algorithm}
\caption{Dynamic Boosted Random Forest}
\label{alg:1}
\begin{algorithmic}[1]
\REQUIRE Training set: $\mathcal{D}^d$, Test set: $\mathcal{D}^t$, Iterations $n$
\ENSURE Prediction of test set: $\mathcal{O}$
\STATE // Training Procedure
\STATE $\mathcal{D} \leftarrow \mathcal{D}^d$
\FOR{$i = 1 \to n$}
    \STATE Train RF as $\mathcal{F}_i$ on dataset $\mathcal{D}$
    \STATE Get scores of leaf nodes $\Pi_i$ by HEM (Section \ref{hem})
    \STATE Split $\mathcal{D}$ into easy data $\mathcal{D}_e$ and hard data $\mathcal{D}_h$ according to $\mathcal{F}_i$ and $\Pi_i$
    \STATE $\mathcal{D} \leftarrow \mathcal{D}_h$
    \STATE add $\mathcal{F}_i$ to $\mathcal{F}$-list
    \STATE add $\Pi_i$ to $\Pi$-list
\ENDFOR
\STATE
\STATE // Test Procedure 
\STATE $\mathcal{D} \leftarrow \mathcal{D}^t$
\STATE $\mathcal{O} \leftarrow \varnothing$
\FOR{$i = 1 \to n$}
    \STATE $\mathcal{F}^d_i \leftarrow \mathcal{F}$-list at $i$
    \STATE $\Pi_i \leftarrow \Pi$-list at $i$
    \STATE Split $\mathcal{D}$ into easy data $\mathcal{D}_e$ and hard data $\mathcal{D}_h$ according to $\mathcal{D}_i$ and $\Pi_i$
    \STATE Predict $\mathcal{D}_e$ as $\mathcal{O}'$ by $\mathcal{F}_i$
    \STATE $\mathcal{D} \leftarrow \mathcal{D}_h$
    \STATE $\mathcal{O} = \mathcal{O} \cup \mathcal{O}'$
\ENDFOR\STATE Predict $\mathcal{D}_h$ as $\mathcal{O}'$ by $\mathcal{F}_n$
\STATE $\mathcal{O} = \mathcal{O} \cup \mathcal{O}'$
\STATE \textbf{return} $\mathcal{O}$
\end{algorithmic}
\end{algorithm}

\section{Proposed method}
\label{propose}
In this section, we present our proposed deep dynamic boosted forest (DDBF), which drives the evolution of the model by iteratively updating the training data. Compared to RF, DDBF can greatly improve the performance. We will first present the general framework and the basic algorithm of our model and then propose two mechanisms to enhance the basic model.

Then, we consider a typical classification task where $X$ and $Y$ denote input and output space respectively. For a decision tree $\mathcal{T}$, $\mathcal{N}$ denotes a decision node and $\mathcal{L}$ a leaf node. $\mathcal{F}$ is a set of $\mathcal{T}$ trained on data $\mathcal{D} = \left\{\mathcal{X}, \mathcal{Y}\right\}$ where $\mathcal{X}$ is point set in $X$ and $\mathcal{Y}$ is corresponding labels set in $Y$.

\subsection{The general framework}
\label{3.1}
Figure \ref{fig:2} illustrates the basic structure of DDBF. We first split a dataset into training set and test set separately and then train a random forest on the training set $\mathcal{D}^d = \left\{\mathcal{X}, \mathcal{Y}\right\}$. Next, we use a criterion to measure the quality of each leaf node of each decision tree in the forest. In Figure \ref{fig:2}, the leaf nodes circled in bold line are good ones which represent possible easy examples. By using the proposed hard example mining (HEM) method to be elaborated in Section \ref{hem}, we can divide $\mathcal{D}^d$ into two parts $\left\{\mathcal{D}{_e^d}, \mathcal{D}{_h^d}\right\}$, where $\mathcal{D}{_e^d}$ denotes easy examples and $\mathcal{D}{_h^d}$ denotes hard examples. Only $\mathcal{D}{_e^d}$ are preserved for the next iterations training. This process keeps on iterating until predetermined $n$ iterations are done.

At the $i^{th}$ iteration, we need to preserve the random forest model $\mathcal{F}_i$ trained in current iteration and the evaluation scores of all leaf nodes of all decision trees in $\mathcal{F}_i$, denoted as $\Pi_L$. For predicting test set $\mathcal{D}^t$, we first use $\mathcal{F}_i$ to predict and then divide $\mathcal{D}^t$ into $\left\{\mathcal{D}{_e^t}, \mathcal{D}{_h^t}\right\}$. For $\mathcal{D}{_e^t}$ the easy data, the predictions made by $\mathcal{F}_i$ are outputted as the final prediction result, while for the hard data $\mathcal{D}{_h^t}$, we will feed them into $\mathcal{F}_{i+1}$ for the next iterations prediction. This process goes on until no longer contains data, or until the last iteration. In the last iteration $n$, the output of $\mathcal{F}_n$ will be the final predicted labels of the corresponding data. This training and test process can be further written in pseudo code shown in Algorithm \ref{alg:1}.

\begin{figure*}
    \centering
    \noindent\makebox[\textwidth]{%
    \includegraphics[width=0.75\textwidth]{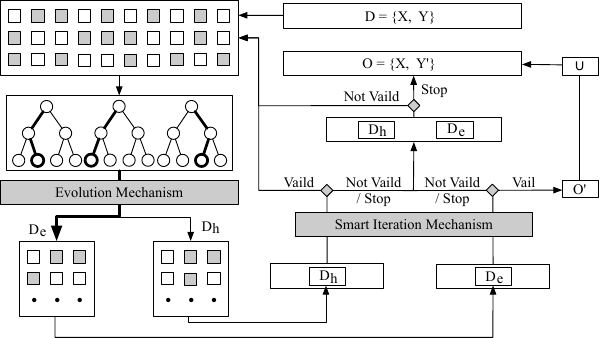}}
    \caption{Illustration of the mechanisms of DDBF. The evolution mechanism is used to eliminate poor decision trees from the random forest and the iteration mechanism is used to better guide the HEM process.}
    \label{fig:3}
\end{figure*}

We propose to use two mechanisms to enhance the basic model of DDBF. To prevent the negative influence of poor decision trees in a random forest on the HEM process, we propose to use an evolution mechanism to eliminate them from the random forest, which will be further elaborated in Section \ref{3.3}. Furthermore, we propose a smart iteration mechanism to better guide the HEM process, as elaborated in Section \ref{3.3}. The model of DDBF with two mechanisms is illustrated in Figure \ref{fig:3}. 

\subsection{Hard example mining (HEM)}
\label{hem}
A random forest consists of multiple decision trees. We use $\mathcal{D}{_e^F}$ and $\mathcal{D}{_h^F}$ to denote easy examples and hard examples of a random forest, $\mathcal{D}{_e^T}$ and $\mathcal{D}{_h^T}$ to denote easy examples and hard examples of each decision tree. We propose that $\mathcal{D}{_e^F}$ and $\mathcal{D}{_h^F}$ can be generated as:

\begin{equation}
\mathcal{D}{_F^e} = \bigcap_{\mathcal{T} \in \mathcal{F}} \mathcal{D}{_T^e}, ~ \mathcal{D}{_F^h} = \mathcal{D} - \mathcal{D}{_F^e}
\end{equation}
meaning that the intersection of easy examples of all the decision trees are considered as easy examples. 

A leaf node $\mathcal{L}$ in a decision tree corresponds to a rule $\mathcal{R_L}$. Suppose the path to reach the leaf node $\mathcal{L}$ from the root in a decision tree $\mathcal{F}$ is $\mathcal{W_L} = \left\{n_1, n_2,\cdots, n_i\right\}$, where $n_1$ to $n_i$ denote the decision nodes along the path. The probability distribution in $\mathcal{Y}$ in node $\mathcal{L}$ is $\pi_L$. $c$ is a label in $\mathcal{Y}$, then we can define the rule $\mathcal{R_L}$ corresponding to the leaf node $\mathcal{L}$ as
\begin{equation}
\mathcal{R_L}:x~|~x \in \bigcap_{n_i \in \mathcal{W_L}} r(n_i) \Rightarrow y = arg_c max~\pi_L (c) \label{eq:2}
\end{equation}
where $r()$ is a function that represents the data that satisfy the rule of a decision node $n$. Then the easy examples of a decision tree can be defined as
\begin{equation}
\mathcal{D}{_T^e} = \{x~|~x \in \bigcap_{\mathcal{L} \in \mathcal{T}, \mathcal{T} \in \mathcal{F}} (score(\mathcal{L}) > \sigma) \}
\end{equation}
where $score()$ is a leaf node evaluation metric and $\sigma$ is the threshold, which is the key design of our model. We provide 2 kinds of threshold: average score and mutual information score. Average score uses the mean value of all leaf nodes while mutual information score is based on information theory. We will discuss the choice of the threshold in Section \ref{conc}.

Next, we propose several leaf node common evaluation metrics $score()$. We may use support and confidence, as were used to evaluate association rules \cite{agrawal1993mining} to score the leaf node. For the rule $\mathcal{R_L}$ of a leaf node $\mathcal{L}$, all data that satisfy the preconditions of the rule, \ie $~x \in r(n_i)$, compose candidate set $\mathcal{C}$. Then
\begin{equation}
score_{supp}(\mathcal{L}) = \frac{|\mathcal{C}|}{|\mathcal{X}|}
\end{equation}

\begin{equation}
score_{conf}(\mathcal{L}) = \frac{|\{y=c, x \in \mathcal{C}\}|}{|\mathcal{C}|}
\end{equation}
Since both support and confidence derive from association rules, we can merge them to be $f1$ score.
\begin{equation}
score_{f1}(\mathcal{L}) = 2 \cdot \frac{score_{supp}(\mathcal{L}) \cdot score_{conf}(\mathcal{L})}{score_{supp}(\mathcal{L}) + score_{conf}(\mathcal{L})}
\end{equation}
Also, since Gini impurity (gini) and information gain (entropy) are the partitioning criteria used in decision tree, we can define gini score and entropy score of a leaf node as 
\begin{equation}
score_{gini}(\mathcal{L}) = - Gini(\mathcal{L}) = \sum_{j=1}^c p_j^2 - 1
\end{equation}

\begin{equation}
score_{entropy}(\mathcal{L}) = - Entropy(\mathcal{L}) = \sum_{j=1}^c p_j^2~log~p_j
\end{equation}
where $c_j$ is $j^{th}$ label in $\mathcal{Y}$ and $p_j$ is the probability of $c_j$.

Above all, we propose 3 leaf node evaluation metrics, namely $score$-$gini$, $score$-$entropy$, and $score$-$f1$. Effects are explored in experiments, as discussed in Section \ref{4.1}.

\subsection{Mechanisms}
\label{3.3}

\begin{table}[]
\caption{Comparison results between DDBF and the other ensemble learning methods on test accuracy with different datasets.}
\label{tab:1}
\centering
\resizebox{\textwidth}{35mm}{
\begin{tabular}{@{}lcccccccccccccc@{}}
\toprule
Dataset   & \multicolumn{2}{c}{ADULT}       & \multicolumn{2}{c}{LETTER}      & \multicolumn{2}{c}{YEAST}       & \multicolumn{2}{c}{SKIN}        & \multicolumn{2}{c}{POKER}       & \multicolumn{2}{c}{MNIST}     & \multicolumn{2}{c}{SATIMAGE}  \\ \midrule
Attribute & \multicolumn{2}{c}{Categorical} & \multicolumn{2}{c}{Categorical} & \multicolumn{2}{c}{Categorical} & \multicolumn{2}{c}{Categorical} & \multicolumn{2}{c}{Categorical} & \multicolumn{2}{c}{Numerical} & \multicolumn{2}{c}{Numerical} \\ \midrule
          & ACC                 & Range     & ACC                 & Range     & ACC                 & Range     & ACC                 & Range     & ACC                 & Range     & ACC                & Range    & ACC                & Range    \\ \midrule
sNDF      & 85.58               & $\pm$0.04      & 97.08               & -         & 60.31               & $\pm$0.02      & 94.21               & $\pm$0.03      & 60.73               & $\pm$0.02      & 96.35              &$\pm$ 0.02     & 91.16              & $\pm$0.02     \\
gcForest  & 86.40               & -         & 97.12               & $\pm$0.03      & 63.45               & -         & 94.35               & $\pm$0.04      & 62.02               & $\pm$0.03      & \textbf{98.25}     & $\pm$0.03     & 91.70              & $\pm$0.03     \\
MLP       & 85.25               & -         & 95.70               & -         & 55.60               & -         & 93.64               & $\pm$0.04      & 58.28               & $\pm$0.02      & 96.62              & $\pm$0.03     & 91.13              & $\pm$0.02     \\
RF        & 85.49               & -         & 96.50               & -         & 61.66               & -         & 92.48               & $\pm$0.02      & 61.83               & $\pm$0.04      & 97.18              & $\pm$0.02     & 91.20              & $\pm$0.02     \\
GBDT      & 86.34               & $\pm$0.02      & 96.32               & $\pm$0.04      & 60.98               & $\pm$0.03      & 94.29               &$\pm$0.04      & 62.58               & $\pm$0.04      & 97.28              & $\pm$0.03     & 89.99              & $\pm$0.03     \\
XGBoost   & 85.90               & $\pm$0.02      & 95.85               & $\pm$0.03      & 59.16               & $\pm$0.03      & 93.80               & $\pm$0.04      & 62.06               & $\pm$0.03      & 97.73              & $\pm$0.02     & 90.45              & $\pm$0.01     \\ \midrule
DDBF-g    & 86.57               & $\pm$0.02      & 97.02               & $\pm$0.03      & 63.68               & $\pm$0.01      & 95.15               & $\pm$0.04      & 63.80               & $\pm$0.05      & 97.98              & $\pm$0.03     & 92.41              & $\pm$0.03     \\
DDBF-e    & 86.56               & $\pm$0.05      & 97.18               & $\pm$0.03      & \textbf{64.13}      & $\pm$0.02      & 95.18               & $\pm$0.01      & 63.87               & $\pm$0.03      & 98.05              & $\pm$0.02     & 92.49              & $\pm$0.02     \\
DDBF-f    & \textbf{86.62}      & $\pm$0.01      & \textbf{97.25}      & $\pm$0.02      & 63.90               & $\pm$0.02      & \textbf{95.21}      & $\pm$0.02      & \textbf{64.01}      & $\pm$0.02      & 98.19              & $\pm$0.02     & \textbf{92.52}     & $\pm$0.01     \\ \bottomrule
\end{tabular}}
\end{table}

\subsubsection{Evolution mechanism}

In RF, data and features are randomly sampled to generate multiple decision trees. Chances are that some of them are of poor quality, which have negative effect on the voting process in our proposed HEM method. To overcome this problem, we draw on the idea of genetic programming \cite{banzhaf1998genetic} and propose to use a fitness formula to eliminate those decision trees with lower scores before determining easy and hard examples from the training data. In our implementation, we use the average evaluation metric score across all leaf nodes as the fitness score of each decision tree. The specific procedure is as follows.
\begin{list}{\labelitemi}{\leftmargin=1em}
    \setlength{\topmargin}{0pt}
    \setlength{\itemsep}{0em}
    \setlength{\parskip}{0pt}
    \setlength{\parsep}{0pt}
\item At each iteration, calculate the fitness score of each tree in a random forest after training.
\item Set an elimination ratio (20$\%$ by default) and calculate the threshold, and then eliminate those decision trees whose score is lower than the threshold.
\item Determine the easy and hard examples by voting among the rest of the decision trees and remove them to generate a new training set for the next iteration.
\end{list}

\subsubsection{Iteration mechanism}

At each iteration during training, the division of $\mathcal{D}^d$ as $\left\{\mathcal{D}{_e^d}, \mathcal{D}{_h^d}\right\}$, might not be ideal in terms of validation accuracy. We propose to use the prediction accuracy on the training data $\mathcal{D}^d$ as the validation accuracy (which can be generated using k-fold cross-validation as mentioned in Stacking \cite{breiman1996stacked,wolpert1992stacked}, to judge the quality of the division and annul the division or terminate training if rules are triggered. We propose two smart iteration rules as follows.
\begin{list}{\labelitemi}{\leftmargin=1em}
    \setlength{\topmargin}{0pt}
    \setlength{\itemsep}{0em}
    \setlength{\parskip}{0pt}
    \setlength{\parsep}{0pt}
\item If the validation accuracy decreases for $N$ consecutive times (five by default), apply early termination.
\item If the validation accuracy of easy examples $\mathcal{D}^e$ is lower than that of the whole training data $\mathcal{D}^d$, render this division invalid and continue to train on $\mathcal{D}^d$ in the next iteration.
\end{list}

\section{Experiments}
To evaluate DDBF, we compared its performance on public datasets against several popular or related methods, and performed ablation test to determine the effects of the three proposed mechanisms for enhancement. We further applied visualization to demonstrate the effectiveness of DDBF.

In these experiments, we used the default parameters defined in our model and did not fine-tune them. The default number of decision trees in a random forest is 200, the default splitting criterion in a decision tree is Gini impurity; the default number of iterations is 10, the default quality evaluation criterion for leaf nodes is $score$-$f1$, the other parameters have already been mentioned in the Section \ref{propose}.

\subsection{Performance}
\label{4.1}
We used seven classification benchmark datasets with different scales. The datasets vary in size: from 1484 up to 78823 instances, from 8 up to 784 features, and from 2 up to 26 classes. We provide five datasets from the UCI Machine Learning Repository in the comparison experiments, namely Adult, Letter, Yeast, Skin and Poker.The former three are balanced and the last two are unbalanced. Except for MNIST \cite{lecun1998gradient} dataset, SATIMAGE dataset is obtained from LIBSVM \cite{chang2011libsvm} dataset. Based on the attribute characteristics of the dataset, we classify the datasets into two categories: categorical, and numerical modeling tasks. We evaluated three versions of DDBF, namely DDBF-g, DDBF-e and DDBF-f, which uses $score$-$gini$, $score$-$entropy$, and $score$-$f1$ as the quality evaluation criterion of leaf nodes respectively. For comparison, we choose two ensemble algorithms, GBDT \cite{friedman2001greedy} and RF \cite{breiman2001random}, two related state-of-the-art approaches, gcForest \cite{zhou1702deep} and sNDF \cite{kontschieder2015deep}, and multiplayer perception (MLP), and XGBoost \cite{chen2016xgboost}. 

For fairer comparison, we compiled the open-source code of gcForest published by Professor Zhou’s team and used exactly the same method to split a dataset into training set and test set. The evaluation metric of the experiments is accuracy, as shown in Table \ref{tab:1}. We used PyTorch to reproduce sNDF and recorded the results on the datasets. We used scikit-learn library to evaluate GBDT and XGBoost. As Table \ref{tab:1} shows, DDBF performs better than RF in all three datasets, which indicates that DDBF is indeed superior to RF. For the multilayer perceptron (MLP) configurations, we use ReLU for the activation function, cross-entropy for the loss function, adadelta for optimization, no dropout for hidden layers according
to the scale of training data. The network structure hyper-parameters, however, could not be fixed across tasks. Therefore, for MLP, we examine a variety of architectures on the validation set, and
pick the one with the best performance, then train the whole network again on the training set and report the test accuracy. Almost all DDBF versions achieve state-of-the-art results, proving evidence to the effectiveness of DDBF. Last but not least, the results in the last three rows indicate that $score$-$f1$ may be the best fit for the quality evaluation criterion of leaf nodes, though not by a large margin over the other two.

\subsection{Decision regions visualization}
\label{4.2}
We visualized the decision boundaries learned by DDBF to verify that our proposed hard example mining method indeed achieves our intended purpose.

\begin{figure*}
    \centering
    \noindent\makebox[\textwidth]{%
    \includegraphics[width=0.8\textwidth]{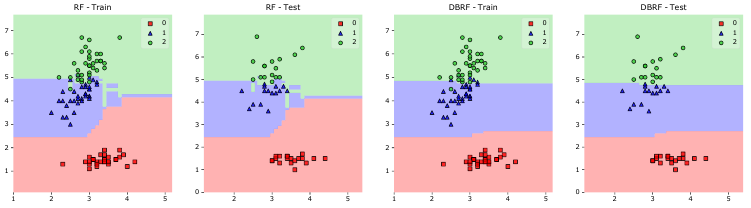}}
    \caption{Visualization of decision boundaries. The two of the training set and test set, while the two pictures the training set and test set, respectively.
}
    \label{fig:5}
\end{figure*}

To facilitate visualization, we selected the second and third column of the UCI Iris dataset, and randomly split the dataset into 67$\%$ for training and 33$\%$ for testing. We conducted a comparison experiment between RF and DDBF using default parameters (for RF we used scikit-learn). RF achieves 98$\%$ accuracy on the training set but only 94$\%$ accuracy on the test set, which indicates overfitting. In contrast, though DDBF achieves 94$\%$ accuracy on the training set, it achieves a surprising 98$\%$ accuracy on the test set, suggesting that it has learned better decision boundaries. The decision boundaries learned by RF and DDBF are shown in Figure \ref{fig:5}, which corroborates the above findings.

\begin{table*}[]
\caption{Comparison of AUC and running time on Credit Card Fraud dataset on 3 data splits.}
\label{tab:4}
\centering
\resizebox{\textwidth}{25mm}{
\begin{tabular}{@{}lcccc@{}}
\toprule
         & \multicolumn{2}{c}{AUC(Test)}                        & \multicolumn{2}{c}{Time(s)}        \\ \cmidrule(l){2-5} 
         & Avg.                          & Std.                 & Avg.              & Std.           \\ \midrule
sNDF     & 0.9348/0.9290/0.9264          & 1.2849/1.3254/1.1094 & 712.4/717.0/724.4 & 20.9/34.0/24.5 \\
gcForest & 0.9580/0.9702/0.9548          & 0.2208/0.3660/0.1673 & 161.0/152.2/148.4 & 8.1/8.1/4.5    \\
MLP      & 0.9172/0.8423/0.9112          & 1.9070/4.4489/1.9399 & 51.7/53.8/62.2    & 9.9/8.7/8.9    \\
RF       & 0.9508/0.9598/0.9478          & 0.3050/0.5299/0.2218 & 61.2/61.6/61.5    & 1.4/1.8/1.4    \\
GBDT     & 0.9434/0.8829/0.7259          & 0.0009/0.0032/0.0063 & 147.4/141.6/135.2 & 8.6/6.7/3.1    \\ \midrule
DDBF     & \textbf{0.9762/0.9848/0.9730} & 0.0293/0.0242/0.1117 & 97.4/81.0/77.8    & 5.7/2.8/0.8    \\ \bottomrule
\end{tabular}}
\end{table*}

\subsection{Imbalanced data}
\label{4.3}
We can better understand how our proposed hard example mining method makes DDBF more effective from a sampling point of view. Suppose we have a large dataset and the decision boundaries are determined by only a small proportion of the dataset. These are important data that ideally need to be preserved for training. However, traditional sampling methods such as random sampling used in RF have no clue whether a data point is important or not. Since they account for only a small amount, chances are that after sampling there are simply not enough important data left for a classifier to learn from. This is how a model may fail to learn the boundaries in spite of seemingly abundant data and features. Removing easy examples makes the proportion of important data higher, and thus makes it more likely that there will be sufficient amount of important data left after sampling. To the best of our knowledge, our proposed hard example mining method is a new way of sampling.

This is extremely useful when dealing with imbalanced datasets. We applied DDBF to MNIST dataset and Kaggle’s classic imbalanced datasets, Credit Card Fraud dataset to demonstrate it.

\begin{figure}
    \centering
    \includegraphics[width=0.65\textwidth]{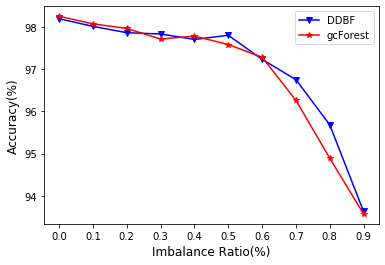} 
    \caption{Influence of imbalanced labeled samples to the classification accuracy on MNIST.
}
    \label{fig:6}
\end{figure}

MNIST handwritten dataset has a training set of 60000 samples and a test set of 10000 samples, each sample has a pixel of 28 $\times$ 28, and each pixel is a gray-scale range from 0 to 255. Since number five is often easily confused with number six, we implement this experiment with imbalanced data to test our performance. We gradually increase the imbalance ratio of class with label 5, by means of reducing the number of samples with label 5 in the training set. The imbalance ratio is equal to the number of dropped samples divided by the used one. Figure \ref{fig:6} shows that the classification accuracy of each algorithm curves as we continue increasing the imbalanced ratio of labeled dataset.

Though our proposed DDBF doesn't outperform gcForest when the labeled dataset MNIST is balanced at the beginning, classification accuracy of gcForest is decreased significantly when the imbalance ratio increases gradually. Figure \ref{fig:6} illustrates that our proposed DDBF performs better on the imbalanced data, which proves the effectiveness of hard examples mining. It also can be seen that when imbalance ratio is beyond 0.6, our DDBF even outperforms gcForest about 0.8$\%$.      

\begin{figure}
    \centering
    \includegraphics[width=0.65\textwidth]{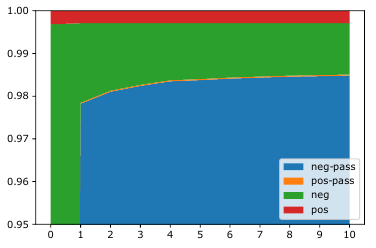} 
    \caption{Stacked training data distribution in each iteration. The blue area (neg-pass) and the yellow area (pos-pass) at the bottom represent the proportions of negative and positive examples that were removed after each iteration. The red area (pos) and green area (neg) at the top represent the proportions of positive and negative examples that were kept as hard examples after each iteration.
}
    \label{fig:4}
\end{figure}

Besides, we randomly selected 65$\%$ of the dataset as training data and 35$\%$ as test data. Figure \ref{fig:4} shows how the proportion of positive and negative examples in the training data shifted after each iteration. We can see that in the initial training set (on the left of the x-axis), the data are extremely imbalanced with a negative-to-positive ratio of 565.38:1. As the iteration goes on, more data are removed from the training set and at the end (on the right of the x-axis), data are more balanced in the remaining hard examples for training with a negative-to-positive ratio of 3.62:1.

The whole process achieves the purpose of hard example mining and sampling at the same time. Essentially, it uses data’s importance rather than its label as a guideline for sampling. Without much fine-tuning, DDBF achieves an AUC 97.55$\%$ on the test set, a very competitive score.

\subsection{Iteration mechanism}
We designed two experiments to verify the effectiveness of the iteration mechanism of DDBF. We first train other methods on the remaining training data after the first iteration of DDBF. Table \ref{tab:2} shows their overall test AUC all increased, which indicates that the iterative mechanism to divide the data has an improved classification effect.
\begin{table*}[]
\caption{Comparison of AUC of other methods trained alone and on top of the first iteration of DDBF.}
\label{tab:2}
\centering
\begin{tabular}{@{}llllll@{}}
\toprule
      & MLP   & RF    & GDBT  & sNDF  & gcForest \\ \midrule
alone & 91.72 & 95.08 & 94.34 & 93.48 & 95.80    \\
+DDBF & 95.98 & 97.40 & 97.38 & 96.69 & 97.37    \\ \bottomrule
\end{tabular}
\end{table*}

We also fixed the number of trees per forest and iterated the DDBF three times, so that RF and DDBF could be better compared with a certain total number of trees. Table \ref{tab:3} records the test AUC of RF and DDBF after 3 iterations when the number of trees in a forest is fixed. It shows that just like neural nets, depth is also beneficial to RF.
\begin{table}[]
\caption{Comparison of AUC between RF and DDBF after three iterations when number of trees per forest is fixed.}
\label{tab:3}
\centering
\begin{tabular}{@{}lcccccc@{}}
\toprule
     & 17    & 50    & 150   & 450   & 1350  & 4050  \\ \midrule
RF   & -     & 93.86 & 94.58 & 96.59 & 96.60 & 96.67 \\
DDBF & 97.21 & 97.25 & 97.57 & 97.62 & 97.83 & -     \\ \bottomrule
\end{tabular}
\end{table}

\subsection{Running time}
The 3 UCI datasets we used for evaluation differ in data size (1.5k, 20k, 50k), ranging from small to medium. To further consolidate our claim that the proposed model DDBF is superior to RF and can achieve state-of-the-art result while still being computational efficient, we conducted a series of more experiments on Kaggle's Credit Card Fraud dataset mentioned in section \ref{4.3}. It has a considerably larger data size of 280k, and is extremely imbalanced since only about 400 data have positive labels. Thus, we have to use AUC instead of accuracy as the evaluation metric. We also reported standard deviation and computation time where necessary.

We compared our DDBF model against several closely related methods in terms of AUC and training time. As the original data are given in time order, we performed shuffling before randomly splitting the data into training set and test set with a ratio of roughly 2:1 ($\%$65:$\%$35). To be more credible, we ran the experiments on 3 different data splits, and for each split we ran each method 5 times and recorded means and standard deviations.

For RF-based models (gcForest, RF and DDBF), we grew 200 trees in a random forest per iteration, considering there are 30 features in the dataset. For iterative models (sNDF, gcForest, DDBF and GBDT), we set iteration times as 10 except for GBDT, for we found that the validation AUC of the first 3 models didn't change much after roughly 10 iterations. And since after 10 iterations, GBDT still performed not well enough so we increased it to 200, when the validation AUC stabilized. All the training time reported were recorded at the end of maximum iterations. For MLP or deep neural network configurations, we used ReLU for activation function, cross-entropy for loss function, Adam for optimization. Via three-fold cross-validation, we set the MLP with 3 hidden layers, 200 neurons in the first layer, 100 in the second and 10 in the third. The other hyper-parameters were set as their default value.

From Table \ref{tab:4} one can see that DDBF performs consistently better than all other methods in terms of AUC on test set on all three data splits. Besides, the training time of DDBF is quite competitive. In particular, it is more computationally efficient than sNDF and gcForest.

We also want to know how the computation time and the size of training data change over each iteration. We recorded computation time and the proportion of easy examples that are removed at each iteration during both training and testing process on the first data split. The results are shown in Table \ref{tab:5}. For comparison, RF trains in 58.2s and costs 0.33s in testing on average.

\section{Discussion}
\subsection{Threshold of hard examples}
As for the hard examples, the threshold for evaluating leaf nodes is very essential. We define hard examples as data that are distributed near the decision boundaries and cannot be classified with high confidence. To efficiently separate hard examples from easy examples, we need a suitable threshold. In this paper, we propose two ways of choosing the threshold: average score and mutual information score.

It is straightforward to exploit the average score of leaf nodes in the decision trees as the threshold. We select hard examples as those whose scores are lower than the mean value in each iteration. This design achieves good results in experiments. However, as for the imbalanced data, the average score may deviate from our desired threshold. Actually, threshold too strict influences the efficiency of our algorithm while loose one may decrease the classification accuracy. Therefore, we seek for more suitable trade-off threshold. 

In \cite{sethi1982hierarchical}, the concept of mutual information is introduced into classification. The amount of of average mutual information obtained about the class set $\mathcal{C}$ from the observation $\mathcal{X}$ can be written as 
\begin{equation}
I(\mathcal{C},\mathcal{X})  = \sum_{i=1}^n\sum_{j=1}^c p(x_i,c_j)\log{\frac{p(c_j|x_i)}{p(c_j)}}
\end{equation}
For better classification, the choice of threshold should be such that we get maximum information about the predictions from inputs. When given $\mathcal{X}$ with imbalanced distribution, those data near decision boundaries is important for the classification results. Therefore, by setting the mutual information as the threshold, we can acquire more knowledge about $\mathcal{Y}$ from those data, and the classification accuracy gets higher. Following this idea, we can filter hard examples really near the decision boundaries instead of depending on the mean value of predictions themselves. Actually, this design may improve accuracy or iteration efficiency compared with the other in experiments as a reasonable trade-off threshold. 

\begin{table}[]
\caption{Computation time and the proportion of easy examples of the first 3 iterations of DDBF on the first data split.}
\label{tab:5}
\centering
\begin{tabular}{@{}lccccc@{}}
\toprule
                           &       & Iter.1  & Iter.2 & Iter.3 & All     \\ \midrule
\multirow{2}{*}{Time(s)}   & Train & 59.18   & 1.05   & 0.91   & 61.14   \\ \cmidrule(l){2-6} 
                           & Test  & 2.86    & 0.16   & 0.16   & 3.18    \\ \midrule
\multirow{2}{*}{Pass rate} & Train & 98.08\% & 0.16\% & 0.12\% & 98.36\% \\ \cmidrule(l){2-6} 
                           & Test  & 98.05\% & 0.15\% & 0.11\% & 98.31\% \\ \bottomrule
\end{tabular}
\end{table}

\section{Related work}
Ensemble learning \cite{dietterich2002ensemble} is a powerful machine learning technique in which multiple learners are trained to solve the same problem. RF \cite{breiman2001random}, GBDT \cite{friedman2001greedy}, XGBoost \cite{chen2016xgboost} are paradigms of ensemble learning, which are all tree-based learning algorithms. DDBF is a novel ensemble algorithm in that it incorporates boosting into the training process of RF. AdaBoost \cite{freund1995desicion} iteratively adds weight to the wrongly classified examples so as to focus training on these hard examples. In contrast, DDBF removes easy examples in each iteration to train on hard examples, which can effectively avoid the effects of unconcerned data while achieving similar boosting effect.

Deep learning research community is also resorting to the strength of tree models. \cite{kontschieder2015deep} proposed deep neural decision forests, which is a novel approach that unifies decision trees with the representation learning known from deep convolutional networks, by training them in an end-to-end manner. In contrary, we attempt to incorporate boosted cascaded structure into deep forests without training through back propagation and hyper-parameter tuning.

DDBF uses RF as the base learner. The capacity of RF is extended vertically by iteratively mining and training on hard examples. \cite{zhou1702deep} proposed gcForest has a cascade procedure similar to DDBF, but the specific approach is different. GcForest cascades the base learner by passing the output of one level of learners as input to the next level, which is similar to Stacking \cite{wolpert1992stacked}. DDBF cascades the base learner by using the quality of leaf nodes in one level as a criterion to dynamically evolve the training data to train the next level (model of the next iteration).

\section{Conclusion}
\label{conc}
In this paper, to improve the performance of Random Forest (RF), we incorporate boosted cascaded structure into the training process of RF and propose a novel deep dynamic boosted forest (DDBF). Specifically, we propose a criterion to measure the quality of a leaf node of all decision trees and then vote to remove easy examples. By iteratively mining and training on hard examples, we evolve the model to learn decision boundaries better and thus extend RF vertically. We also propose evolution mechanism and smart iteration mechanism to enhance DDBF. Experiments show that DDBF outperforms RF and achieves on-par, if not better, results compared to state-of-the- art methods on datasets of various size. And the effectiveness of DDBF can be best shown when applied to imbalanced datasets. We also provide an explanation of its effectiveness from a sampling point of view. We believe that DDBF is a very practical approach and is particularly useful in learning imbalanced data.

\section*{Acknowledgement}
Thanks for the constructive comments of all the reviewers sincerely. We are also grateful to all the people who join the discussion and provide their suggestions. 

\bibliography{acml20}

\end{document}